%% file: main.tex
\documentclass[11pt]{article}
\usepackage[preprint]{acl}

\usepackage{times}
\usepackage{latexsym}
\usepackage[T1]{fontenc}
\usepackage[utf8]{inputenc}
\usepackage{microtype}
\usepackage{graphicx}
\usepackage{booktabs}
\usepackage{multirow}
\usepackage{array}
\usepackage{xspace}
\usepackage{amsmath}
\usepackage{amssymb}
\usepackage{url}
\usepackage{hyperref}
\graphicspath{{./}}

\newcommand{\TE}{\ensuremath{\mathrm{TE}}\xspace}
\newcommand{\B}{\TE}                                 
\newcommand{\LLMLtwo}{\ensuremath{\text{LLMLingua-2}}\xspace}
\newcommand{\MuSiQue}{MuSiQue\xspace}
\newcommand{\TwoWiki}{2Wiki\xspace}
\newcommand{\HotpotQA}{HotpotQA\xspace}

\newcommand{\hopc}{\ensuremath{\text{hop}_c}\xspace}
\newcommand{\Cprime}{\ensuremath{\text{C1}'}\xspace}

\newif\ifsubmitmode
\submitmodetrue

\title{Context Compression Is Not One Thing:\\
       Readable Symbolic Re-expression vs.\ Coherent Summary at Matched Budget}

\author{
Sisong Bei \\
Independent Researcher \\
\texttt{qurining@gmail.com} \And
Mikhail L. Arbuzov \\
Independent Researcher \\
\texttt{mike.arbuzov54@gmail.com} \AND
Ziwei Dong \\
Independent Researcher \\
\texttt{ziwei.dong@alumni.emory.edu} \And
Dmitri Kalaev \\
Independent Researcher \\
\texttt{kalaevdr@gmail.com} \AND
Alexey Shvets \\
Palo Alto Networks \\
\texttt{ashvets@paloaltonetworks.com}
}

\begin{document}
\maketitle

\input{0_abstract}
\input{1_introduction}
\input{2_related_work}
\input{3_method}
\input{4_experiments}
\input{5_results}
\input{6_discussion}
\input{8_conclusion}
\input{7_limitations}

\bibliography{references}

\appendix
\input{A_appendix}

\end{document}

%% file: 0_abstract.tex
\begin{abstract}
We study context compression for multi-hop question answering with small
language models. We propose Telegraph English, a readable symbolic format
that rewrites retrieved passages into structured entity-relation
statements, preserving reasoning evidence at lower token cost. In
controlled experiments on \MuSiQue, \TwoWiki, and \HotpotQA, Telegraph
English outperforms three matched-budget compression baselines
(character-level deletion, truncation, and random subsampling) on every
dataset, with gains of 13 to 20 F1 points. It also outperforms a coherent
prose summary produced by the same encoder on the hardest dataset. A
pre-registered depth-interaction hypothesis is null: the advantage does
not grow with reasoning depth within datasets. We interpret these results
as evidence that readable symbolic re-expression preserves entity content
more densely than either natural language or coherent summarization at
matched token budget.
\end{abstract}

%% file: 1_introduction.tex
\section{Introduction}
\label{sec:intro}

Small language models face a tension on multi-hop question answering:
retrieved natural-language (NL) context is expensive in tokens and
error-prone in long passages, while short retrieval discards the bridge
entities a reader needs for multi-step reasoning. Prior context-compression
work attacks this tension through token-level scoring
\citep{jiang2023llmlingua,jiang2024llmlingua2}, hidden-state
summarisation \citep{mu2024gist,chevalier2023autocompressor}, or
task-aware abstractive summarisation \citep{xu2024recomp}, but each
commits to selective retention or latent-space summary rather than
re-expression in surface text.

We study a different move. A small encoder rewrites the retrieved NL
passage into a readable, rule-governed symbolic format we call
\emph{Telegraph English} (TE), and a small consumer reads TE in place of
NL. TE preserves entities verbatim and replaces connective NL tissue
with pipe-separated symbolic operators. The encoder prompt is fixed and
the encoder is frozen; the consumer needs no fine-tuning.

This work builds on Telegraph English, a readable symbolic compression
format introduced in \citet{arbuzov2026semantic}, and is motivated by
the error-accumulation analysis in \citet{arbuzov2025error} showing that
LLM errors concentrate at a sparse subset of key tokens. The theoretical
framework for patch-local reliability engineering
\citep{arbuzov2026architecture} provides the broader context.

We test TE on three multi-hop benchmarks (\MuSiQue, \TwoWiki,
\HotpotQA) against three matched-budget controls: character-level
density matching, end-truncation, and random subsampling. TE wins on
every dataset and every control, with paired-bootstrap 95\% confidence
intervals strictly positive and gains of $+13.6$ to $+20.2$ F1 points
(Fig.~\ref{fig:mechanism}). TE also beats a coherent prose summary
produced by the same encoder at the same token budget on the hardest
dataset (\MuSiQue, $+11.94$ pp). The matched-budget controls rule out
character-density manipulation, NL-tail dispensability, and random-token
sufficiency as alternative explanations for TE's advantage.

We pre-registered a stronger depth-interaction hypothesis: that TE's
advantage over NL would grow with the reasoning depth of the question.
This hypothesis is null. All four within-dataset interaction slopes are
direction-consistent with the prediction but none is statistically
significant (FDR-corrected $p > 0.41$, $I^2 = 0\%$). A minimum-detectable-
effect-size analysis bounds the design to ruling out widenings of roughly
$4$--$5$ F1 points across the hop-2-to-hop-4 range; weaker effects cannot
be distinguished from zero at our sample size.

\paragraph{Contributions.}
\begin{itemize}
  \setlength\itemsep{0.15em}
  \item A matched-budget compression mechanism for multi-hop QA: at the
    same token budget, TE beats three trivial-compression controls and a
    coherent-prose summariser, supporting the reading that TE compresses
    where natural language carries redundancy.
  \item A pre-registered null on depth-dependent widening, paired with a
    minimum-detectable-effect-size analysis that bounds the scope of the
    matched-budget gains to constant offsets rather than depth-scaling
    advantage.
\end{itemize}

%% file: 2_related_work.tex
\section{Related Work}
\label{sec:related}

Context compression for language models has emerged as a response to the
growing cost of long retrieval-augmented contexts. Our contribution
differs from prior work in mechanism: TE is a surface-text re-expression
produced by a frozen encoder with a fixed prompt, read by a frozen
consumer. We organise prior work into four families and contrast TE with
each.

\paragraph{Token-level scoring.}
LLMLingua \citep{jiang2023llmlingua} and LLMLingua-2
\citep{jiang2024llmlingua2} score individual tokens for retention or
deletion via a small model trained on information-preservation proxies.
LongLLMLingua \citep{jiang2024longllmlingua} extends scoring to
document-level saliency. Our primary baseline of this family is
\LLMLtwo at rate-50. The mechanism is selective retention: the output is
a subsequence of the input. TE's mechanism is re-expression: the encoder
rewrites content into an entity-preserving format in which bridge
entities are kept verbatim and connective tissue is replaced by
symbolic operators. Token-level scoring cannot produce that
reformatting at matched budget.

\paragraph{Hidden-state compression.}
GIST \citep{mu2024gist} compresses context into soft tokens at the
hidden-state layer. AutoCompressor
\citep{chevalier2023autocompressor} compresses long contexts into
summary vectors. CEPE \citep{yen2024longcontext} extends cross-attention
to a compressed summary of retrieved passages. All three operate in the
consumer's latent space and require consumer-side training. TE operates
in surface text and needs no consumer-side training---a practical
distinction for small-model deployment where consumer-side retraining
is costly and auditability matters.

\paragraph{Task-aware abstractive summarisation.}
RECOMP \citep{xu2024recomp} compresses retrieved passages via a learned
abstractive summariser fine-tuned on the downstream QA task. CompAct
\citep{yoon2024compact} uses a task-conditioned encoder tuned on QA
supervision. The encoder's output is natural language and the encoder
is fine-tuned on the task. TE's encoder is task-agnostic, runs a frozen
prompt, and produces symbolic-structural output. The contrast is again
mechanism rather than ratio.

\paragraph{Symbolic intermediates for reasoning.}
Prior work has used symbolic intermediates in the forward direction:
chain-of-thought \citep{wei2022cot}, program-aided language models
\citep{gao2023pal}, structured scratchpads \citep{nye2021scratchpad},
where the consumer produces symbolic output. We use a symbolic
intermediate in the input direction: the consumer reads symbolic
context in lieu of NL prose. The Telegraph English format itself was
introduced by \citet{arbuzov2026semantic} as a structured symbolic
rewriting target for prompt compression; the present paper evaluates
it as a tokenizer-aware, encoder-produced context representation
against matched-budget controls on multi-hop QA.

\paragraph{Positioning.}
TE is re-expression, not retention or latent compression. The matched-
budget controls in this paper isolate the re-expression mechanism from
three trivial alternatives, and the coherent-prose comparator further
isolates it from generic abstractive summarisation at the same budget.

%% file: 3_method.tex
\section{Method}
\label{sec:method}

\subsection{Telegraph English}
\label{sec:method-te}

Telegraph English (TE) is a context representation produced by an
encoder language model (Claude Sonnet~4.6 via AWS Bedrock batch
inference) with a fixed, task-agnostic prompt. The encoder rewrites the
retrieved NL passage into a sequence of pipe-separated symbolic clauses
in which entities are preserved verbatim and connective NL tissue is
replaced by short \texttt{@}-prefixed operators. The prompt instructs
output that is compatible with the consumer model's tokenizer
(Qwen-3.5-9B) so that token budget at the consumer matches what is
written. A representative pre/post pair from \MuSiQue:

\begin{quote}\small\ttfamily
NL: ``Barack Obama was born in Honolulu, Hawaii. Honolulu is the
capital of the state of Hawaii. Hawaii is a state in the United
States.''\\[2pt]
TE: ``Barack Obama @born Honolulu | Honolulu @capital\_of Hawaii |
Hawaii @state\_in United States.''
\end{quote}

The full encoder prompt and additional examples are in
Appendix~\ref{app:beta-prompt}.

\subsection{Primary hypothesis: depth interaction}
\label{sec:method-primary}

We pre-registered (Appendix~\ref{app:prereg-protocol}) a primary
hypothesis that TE's advantage over NL grows with the reasoning depth
of the question. Operationally, we model question-level correctness as
a function of representation (NL, TE, or \LLMLtwo), centered hop count,
and their interaction, fit per dataset as a binomial generalised
linear model with cluster-robust standard errors by question. The
primary test is the sign and significance of the
\texttt{representation}~$\times$~\texttt{hop\_count} interaction for
TE: a negative slope means TE's edge over NL grows with hop count.
We pool per-dataset slopes across \MuSiQue and \TwoWiki by random-
effects meta-analysis and apply false-discovery-rate correction
across the primary interaction family. \HotpotQA is excluded from
the regression because all its questions are 2-hop. Full estimating
equations, the heterogeneity branch rule, and the random-effects
specification are in Appendix~\ref{app:stat-protocol}.

\subsection{Auxiliary mechanism: matched-budget controls}
\label{sec:method-aux}

We separately pre-registered an auxiliary mechanism observation
(Appendix~\ref{app:prereg-protocol}): TE performs semantic-preserving
compression only where natural language carries redundancy to strip.
The test is whether TE outperforms three trivial-compression controls
at matched per-row token budget. Each control is computed against TE's
per-row qwen-token count and rules out a specific alternative
explanation:

\begin{itemize}
  \setlength\itemsep{0.15em}
  \item \textbf{Character-density.} The NL passage is rescaled at the
    character level so that its qwen-token footprint matches TE's.
    Rules out the hypothesis that TE's gain comes from per-row
    character-density manipulation.
  \item \textbf{End-truncation.} The NL passage is truncated from the
    end at the qwen-token boundary so it has TE's per-row token count.
    Rules out the hypothesis that the NL tail is dispensable.
  \item \textbf{Random subsampling.} A fixed-seed uniform subsample of
    qwen-token positions is drawn from the NL passage, sized to TE's
    budget. Rules out the hypothesis that any random subset of NL
    tokens would suffice.
\end{itemize}

The direction of the controls is pre-registered (TE should beat all
three); the descriptive ``9 of 9 confidence intervals strictly
positive'' summary is post-hoc. Full per-row specifications are in
Appendix~\ref{app:a6a7a8-spec}.

\subsection{Coherent-prose comparator}
\label{sec:method-prose}

A natural follow-up question is whether TE's advantage holds against a
coherent-prose summary at the same budget rather than against trivial
controls. We run the same encoder under a free-prose summary prompt
and post-truncate each summary to TE's per-row token budget. This
matched-budget contrast isolates representation format from compression
ratio: encoder, consumer, and budget are held fixed; only the surface
form of the compressed passage changes.

%% file: 4_experiments.tex
\section{Experimental Setup}
\label{sec:experiments}

\subsection{Datasets and consumer model}

We evaluate on three multi-hop QA benchmarks with distinct depth
profiles: \MuSiQue \citep{trivedi2022musique} ($n = 2{,}417$ questions;
hops $\in \{2,3,4\}$), \TwoWiki \citep{ho2020wiki2} ($n = 1{,}500$;
balanced 500 per hop level), and \HotpotQA \citep{yang2018hotpotqa}
($n = 1{,}000$; all 2-hop). The consumer is Qwen-3.5-9B (base model;
HF eager bf16, greedy decoding) throughout. The encoder for TE is
Claude Sonnet~4.6 via AWS Bedrock batch inference. Retrieved passages
are the gold-plus-distractor contexts released with each benchmark; NL,
TE, and \LLMLtwo all read the same passage set per question.

\subsection{Prompt and answer extraction}

All representations share a single neutral consumer prompt
(Appendix~\ref{app:prompt}). The answer is extracted from the
consumer's final-line output by regex and evaluated against the gold
answer with token-level F1, with binary correctness at $F1 \geq 0.5$.

\subsection{Baselines}

\begin{itemize}
  \setlength\itemsep{0.15em}
  \item \textbf{NL.} Full retrieved passages, unmodified. The standard
    no-compression baseline.
  \item \textbf{\LLMLtwo at rate-50}~\citep{jiang2024llmlingua2}. The
    primary learned-token-scoring baseline.
  \item \textbf{Three matched-budget controls} (character-density,
    end-truncation, random subsampling), each sized to TE's per-row
    qwen-token count. Defined in \S\ref{sec:method-aux} and
    Appendix~\ref{app:a6a7a8-spec}.
  \item \textbf{Coherent-prose summary} produced by the same encoder
    and post-truncated to TE's per-row token budget
    (\S\ref{sec:method-prose}).
\end{itemize}

\subsection{Statistical protocol}

We use paired-bootstrap 95\% confidence intervals over questions
($n_{\text{boot}} = 10{,}000$, seed 0) for landmark and matched-budget
contrasts. The pre-registered depth-interaction tests use a binomial
GLM with cluster-robust standard errors by question, pooled across
datasets by random-effects meta-analysis, with false-discovery-rate
correction across the primary interaction family
\citep{benjamini1995controlling,dersimonian1986meta}. Full statistical
specifications, the pre-committed heterogeneity rule, and the
sensitivity check against a random-intercept fit are in
Appendix~\ref{app:stat-protocol}.

\subsection{Pre-registered protocol}
\label{sec:experiments-prereg}

We filed a pre-registration prior to data collection. The primary
depth-interaction hypothesis, the matched-budget mechanism observation,
the FDR correction family, the random-effects pooling specification,
the heterogeneity branch rule, and the landmark kill-gate tests are
all pre-registered; the protocol and amendment chain are reproduced in
Appendix~\ref{app:prereg-protocol}.

\subsection{Cross-architecture replication}

To check that the matched-budget mechanism is not specific to one
consumer family, we re-run the three matched-budget controls on a
second consumer, Mistral-7B-Instruct-v0.3, on \MuSiQue. Long-context
NL rows do not fit at 24\,GB on this consumer, so the cross-architecture
claim rests on the matched-budget controls (where TE and the controls
have similar lengths) rather than on the full-NL landmark; details in
\S\ref{sec:disc-mechanism}.

%% file: 5_results.tex
\section{Results}
\label{sec:results}

\subsection{Matched-budget mechanism: TE beats every trivial control}
\label{sec:results-mechanism}

\begin{figure}[t]
    \centering
    \includegraphics[width=\columnwidth]{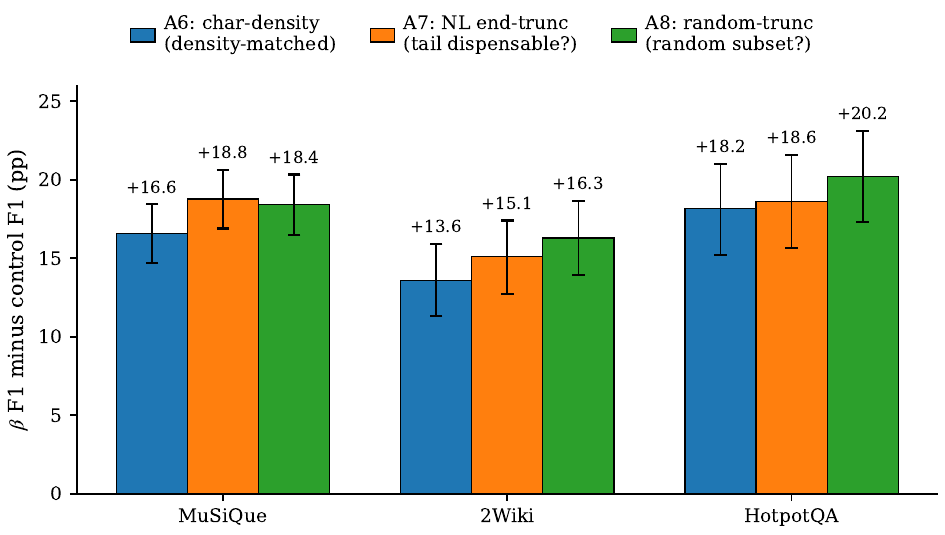}
    \caption{\textbf{Matched-budget mechanism.} TE beats character-density,
    end-truncation, and random-subsampling controls on every dataset at
    matched per-row token budget. Error bars: 95\% paired-bootstrap CI.}
    \label{fig:mechanism}
\end{figure}

Table~\ref{tab:mechanism} reports paired-bootstrap 95\% confidence
intervals for TE versus the three matched-budget controls on each of
\MuSiQue, \TwoWiki, and \HotpotQA. All nine intervals exclude zero from
above, with point estimates ranging from $+13.6$ to $+20.2$ percentage
points (Fig.~\ref{fig:mechanism}).

\begin{table}[t]
\centering\small
\begin{tabular}{llc}
\toprule
Control & Dataset & $\TE - \text{ctrl}$ (pp) [95\% CI] \\
\midrule
Char-density   & \MuSiQue  & $+16.6$ [$+14.7$, $+18.4$] \\
Char-density   & \TwoWiki  & $+13.6$ [$+11.3$, $+15.9$] \\
Char-density   & \HotpotQA & $+18.2$ [$+15.2$, $+21.0$] \\
End-truncation & \MuSiQue  & $+18.8$ [$+16.9$, $+20.6$] \\
End-truncation & \TwoWiki  & $+15.1$ [$+12.7$, $+17.4$] \\
End-truncation & \HotpotQA & $+18.6$ [$+15.7$, $+21.6$] \\
Random-subset  & \MuSiQue  & $+18.4$ [$+16.5$, $+20.3$] \\
Random-subset  & \TwoWiki  & $+16.3$ [$+13.9$, $+18.6$] \\
Random-subset  & \HotpotQA & $+20.2$ [$+17.3$, $+23.1$] \\
\bottomrule
\end{tabular}
\caption{Matched-budget mechanism. At matched per-row qwen-token
budget, TE outperforms all three trivial controls on all three
datasets. Paired bootstrap over questions,
$n_{\text{boot}} = 10{,}000$, seed 0.}
\label{tab:mechanism}
\end{table}

The three controls jointly rule out three specific alternative
explanations for TE's matched-budget advantage. Character-density rules
out per-row character manipulation. End-truncation rules out the
hypothesis that the NL tail is dispensable. Random subsampling rules
out the hypothesis that any size-matched random subset of NL tokens
would suffice. We read the result as supporting the pre-registered
mechanism: TE compresses where natural language carries redundancy,
and trivial alternatives that strip surface tokens but do not
re-express content lose the bridge entities a multi-hop reader needs.

\subsection{Landmark comparisons at full budget}
\label{sec:results-landmark}

Table~\ref{tab:landmark} reports paired-bootstrap 95\% CIs for TE
against full-budget NL and against \LLMLtwo at rate-50. On \MuSiQue,
the deepest-hop dataset, TE beats NL by $+4.75$ pp and \LLMLtwo by
$+7.67$ pp (both $p < 0.001$). On \TwoWiki the TE--NL gap is $+1.53$
pp with a CI that narrowly spans zero. On \HotpotQA the sign flips:
TE loses to NL by $-2.34$ pp ($p < 0.01$).

\begin{table}[t]
\centering\footnotesize
\setlength{\tabcolsep}{4pt}
\resizebox{\columnwidth}{!}{%
\begin{tabular}{lcc}
\toprule
Dataset & $\TE{-}\text{NL}$ (pp) [95\% CI] & $\TE{-}\LLMLtwo$ (pp) [95\% CI] \\
\midrule
\MuSiQue  & $+4.75$ [$+3.06$, $+6.41$]  & $+7.67$ [$+6.00$, $+9.32$] \\
\TwoWiki  & $+1.53$ [$-0.17$, $+3.24$]  & $+1.60$ [$-0.13$, $+3.36$] \\
\HotpotQA & $-2.34$ [$-4.41$, $-0.34$]  & $-2.89$ [$-4.82$, $-0.92$] \\
\bottomrule
\end{tabular}%
}
\caption{Landmark paired comparisons. Full-NL and \LLMLtwo rate-50
budget regimes. Paired bootstrap over questions,
$n_{\text{boot}} = 10{,}000$, seed 0. Per-dataset $n$:
\MuSiQue~2{,}417; \TwoWiki~1{,}500; \HotpotQA~1{,}000.}
\label{tab:landmark}
\end{table}

The between-dataset ordering---positive significant on \MuSiQue,
positive null on \TwoWiki, negative significant on \HotpotQA---is the
empirical pattern that motivates the post-hoc moderator analysis in
Appendix~\ref{app:moderator}; we report it there because at $n = 3$
datasets we cannot rule out unobserved dataset-construction factors.

\subsection{Coherent-prose comparator at matched budget}
\label{sec:results-f3}

A natural concern with the matched-budget mechanism is whether TE's
advantage persists against a coherent-prose summary at the same budget,
rather than against trivial controls that corrupt surface structure.
We run Claude Sonnet~4.6 as a coherent-summary encoder on the same
three datasets and post-truncate each summary to TE's per-row qwen-
token budget. Table~\ref{tab:f3-landmark} reports the comparison.

\begin{table}[t]
\centering\footnotesize
\setlength{\tabcolsep}{4pt}
\resizebox{\columnwidth}{!}{%
\begin{tabular}{lccc}
\toprule
Dataset & $\TE{-}\text{prose}$ (pp) [95\% CI] & $\text{prose}{-}\text{NL}$ (pp) [95\% CI] & $\text{prose}{-}\LLMLtwo$ (pp) [95\% CI] \\
\midrule
\MuSiQue  & $+11.94$ [$+10.09$, $+13.72$]  & $-6.71$ [$-8.58$, $-4.85$]  & $-4.28$ [$-6.13$, $-2.43$] \\
\TwoWiki  & $+0.32$ [$-1.69$, $+2.30$]    & $+1.21$ [$-0.80$, $+3.23$]    & $+1.28$ [$-0.77$, $+3.33$] \\
\HotpotQA & $+1.96$ [$-0.28$, $+4.37$] & $-4.30$ [$-6.60$, $-2.12$] & $-4.85$ [$-7.09$, $-2.69$] \\
\bottomrule
\end{tabular}%
}
\caption{Coherent-prose comparator at matched per-row qwen-token
budget. Paired bootstrap over questions, $n_{\text{boot}} = 10{,}000$,
seed 0. Per-dataset $n$: \MuSiQue~2{,}417; \TwoWiki~1{,}500;
\HotpotQA~1{,}000.}
\label{tab:f3-landmark}
\end{table}

On \MuSiQue, the deepest-hop dataset, TE outperforms the matched-budget
coherent prose by $+11.94$ pp with the CI strictly positive. On
\TwoWiki and \HotpotQA the difference is null. Coherent prose itself
loses to full NL on \MuSiQue and \HotpotQA, suggesting that
matched-budget truncation of coherent summaries is a costly operation
when the budget is tight: at \MuSiQue's budget, the truncated summary
retains roughly half of TE's named entities, while the untruncated
summary at $1.59\times$ the budget retains $78\%$
(Appendix~\ref{app:f3}).

\subsection{Depth interaction is null}
\label{sec:results-null}

We fit the pre-registered depth-interaction model per dataset, pool
\MuSiQue and \TwoWiki by random-effects meta-analysis, and apply
false-discovery-rate correction across the family.
Table~\ref{tab:c1prime} reports the four within-dataset slopes plus
their meta-analytic pool. None is statistically significant: FDR-adjusted
$p$-values range $0.41$ to $0.92$ and the pooled meta slopes are
indistinguishable from zero ($I^2 = 0\%$ on both interaction terms,
so the pre-committed heterogeneity rule did not fire). All four point
estimates are negative---direction-consistent with the pre-registered
prediction---but at these $p$-values the direction-consistency is
indistinguishable from noise, not weak supporting evidence.
Section~\ref{sec:disc-null} quantifies this with a minimum-detectable-
effect-size analysis.

\begin{table}[t]
\centering\footnotesize
\setlength{\tabcolsep}{4pt}
\resizebox{\columnwidth}{!}{%
\begin{tabular}{lcc}
\toprule
Scope & $\TE{:}\text{hop\_c}$ est [95\% CI] & $p_{\text{FDR}}$ \\
\midrule
\MuSiQue ($n{=}2{,}417$) & $-0.018$ [$-0.113$, $+0.077$] & $0.711$ \\
\TwoWiki ($n{=}1{,}500$) & $-0.019$ [$-0.113$, $+0.075$] & $0.711$ \\
Meta (RE)              & $-0.018$ [$-0.085$, $+0.048$] & $0.711$ \\
\midrule
Scope & $\LLMLtwo{:}\text{hop\_c}$ est [95\% CI] & $p_{\text{FDR}}$ \\
\midrule
\MuSiQue & $-0.064$ [$-0.149$, $+0.021$] & $0.413$ \\
\TwoWiki & $-0.004$ [$-0.085$, $+0.077$] & $0.918$ \\
Meta     & $-0.033$ [$-0.091$, $+0.026$] & $0.413$ \\
\bottomrule
\end{tabular}%
}
\caption{Depth-interaction regression. All four within-dataset slopes
are direction-consistent with the pre-registered negative prediction
but non-significant after FDR correction; pooled estimates are
indistinguishable from zero, $I^2 = 0\%$.}
\label{tab:c1prime}
\end{table}

\begin{figure}[t]
    \centering
    \includegraphics[width=\linewidth]{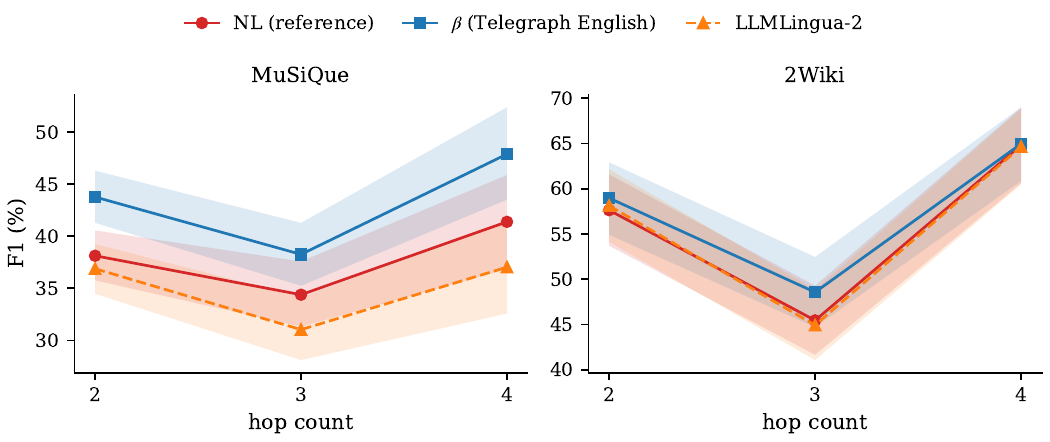}
    \caption{\textbf{Depth-interaction null.} Per-hop F1 for NL, TE, and
    \LLMLtwo within \MuSiQue and \TwoWiki, with 95\% bootstrap CI
    shading. The TE--NL gap is flat across hop counts, not growing with
    depth. \HotpotQA is excluded because all questions are 2-hop.}
    \label{fig:null}
\end{figure}

Figure~\ref{fig:null} visualises the null: the TE--NL gap is flat or
non-monotone across hop levels within both \MuSiQue and \TwoWiki, not
rising with depth as the pre-registered mechanism would predict.

\subsection{Methodological note}
\label{sec:results-methodological}

Our pilot swept three consumer-prompt variants. Under one variant
(role-prompted), TE's \MuSiQue F1 varied by tens of points across
seeds relative to the default prompt because verbose responses interact
with the normalised-multiset F1 scorer in a way that is orthogonal to
context compression. Main-body numbers use the neutral default prompt
throughout; the full prompt-by-metric table is in
Appendix~\ref{app:prompt-metric}.

%% file: 6_discussion.tex
\section{Discussion}
\label{sec:discussion}

\subsection{What the matched-budget result shows}
\label{sec:disc-mechanism}

The matched-budget mechanism result is the central finding of this
paper. At the same per-row qwen-token budget, TE outperforms
character-density, end-truncation, and random-subsampling controls by
$+13.6$ to $+20.2$ percentage points across three datasets. The three
controls exhaust three mechanistically distinct trivial-compression
strategies---compress by character deletion, drop the NL tail, drop
random NL tokens---and none preserves the bridge entities a multi-hop
reader needs. TE preserves bridge entities verbatim and replaces
connective NL tissue with symbolic operators, a re-expression of the
same semantic content at the same budget. We read this as evidence
that TE compresses where natural language carries redundancy: the
matched-budget gain is the benefit of that operation rather than an
artefact of the compression strategies the controls implement.

The coherent-prose comparator sharpens the reading. Coherent prose at
matched budget retains roughly half of TE's named entities on
\MuSiQue, while a longer untruncated coherent summary at $1.59\times$
the budget retains $78\%$. The lost entities are bridge entities the
matched-budget truncation drops, and that loss is a property of
coherent prose at tight budgets, not an artefact of the encoder. TE
holds entity content more densely than either NL or coherent prose
because pipe-separated triples reserve every token for entity-bearing
content that articles, auxiliaries, and connectives would otherwise
consume. We name this the density argument: at matched budget, the
relevant axis is entities-per-token, and TE's representation is the
denser one.

The matched-budget mechanism replicates on a second consumer family.
Re-running the three controls on Mistral-7B-Instruct-v0.3 on \MuSiQue
yields paired-bootstrap 95\% CIs strictly positive on all three
controls ($+8.62 / +11.08 / +10.62$ pp), direction-consistent with
Qwen-3.5-9B at the smaller magnitude expected for a 7B consumer. The
full-NL landmark is not a valid baseline on Mistral because long-context
NL rows do not fit at 24\,GB and the surviving subset is biased toward
shorter (easier) questions; the cross-architecture claim therefore
rests on the matched-budget controls only.

\subsection{Why the null is informative}
\label{sec:disc-null}

The depth-interaction hypothesis predicted that TE's advantage over NL
would grow with hop count within multi-hop datasets. All four within-
dataset slopes came out direction-consistent but non-significant, and
the pooled estimates are indistinguishable from zero. Two readings are
available. The depth-dependent effect may exist but our sample is too
small to detect it. Or the depth-dependent mechanism may simply be
absent: TE's advantage may be a constant offset over compressed-NL
equivalents rather than a depth-dependent widening over full NL.

A minimum-detectable-effect-size analysis bounds the interpretation.
The pooled meta slope is $-0.018$ log-odds per hop with cluster-robust
standard error $0.034$; at $80\%$ power and $\alpha = 0.05$ two-sided,
the minimum detectable slope is roughly $0.095$ log-odds per hop.
Translated to F1 at our consumer's \MuSiQue baseline, that corresponds
to a TE--NL widening of about $4$--$5$ percentage points across the
hop-2-to-hop-4 range. The design therefore rules out widenings of that
magnitude with $\approx\!80\%$ power but cannot distinguish smaller
effects from no effect. The null is informative against a strong-form
depth-dependent advantage and uninformative about a weak-form one.
What our data do support is that the TE--NL gap does not grow
monotonically with hop count within \MuSiQue or \TwoWiki at any
appreciable magnitude (Fig.~\ref{fig:null}).

\subsection{Implications and encoder provenance}
\label{sec:disc-rag}

Our experiments use Claude Sonnet~4.6 as the TE encoder, deliberately
chosen as a strong frontier model so that the consumer-reading question
is not confounded with translator capacity. A natural follow-up is
whether a much smaller encoder, given domain-matched training, could
substitute. A pilot we describe in Appendix~\ref{app:pilot} fine-tunes
Qwen-3.5-0.8B on $5{,}000$ entity-preserving rows with oracle bridge-
entity conditioning, evaluates on the held-out \MuSiQue test shard, and
beats both \LLMLtwo and the Sonnet TE encoder at roughly $10\%$ of the
qwen-token budget. The pilot is single-dataset and uses oracle
conditioning, so it is an upper bound on encoder substitutability
rather than a deployable alternative; the general translator-capacity
question remains open.

%% file: 8_conclusion.tex
\section{Conclusion}
\label{sec:conclusion}

We tested Telegraph English, a readable symbolic re-expression of
retrieved passages, as a matched-budget alternative to natural language
for multi-hop question answering with small language models. Across
three benchmarks, TE outperforms three trivial-compression controls and
a coherent-prose summariser at the same per-row token budget, with all
nine matched-budget confidence intervals strictly positive. A pre-
registered depth-interaction hypothesis is null: the advantage does not
grow with reasoning depth, and a minimum-detectable-effect-size analysis
bounds the design to ruling out widenings of $4$--$5$ F1 points across
hop levels. We interpret these results as evidence that the operative
property of TE is entity density per token, and the natural follow-up
is whether smaller encoders, evaluated without oracle entity
conditioning, can preserve that density.

%% file: 7_limitations.tex
\section{Limitations and Broader Impact}
\label{sec:limits}

\subsection{Limitations}
\label{sec:limits-limits}

\paragraph{Between-dataset gradient is $n=3$.}
The post-hoc reading of the between-dataset TE--NL gradient as
correlating with dataset NL ceiling rests on three datasets. NL-ceiling-
proximity is the only monotone candidate moderator at this $n$;
unobserved dataset-construction factors (bridge-entity retrievability,
distractor-passage content) correlated with NL ceiling cannot be ruled
out. A confirmatory replication would require a fourth dataset where NL
ceiling varies independently of dataset construction, and we report the
gradient as an exploratory observation only
(Appendix~\ref{app:moderator}).

\paragraph{Cross-architecture coverage is mechanism-only.}
The matched-budget mechanism is tested on two consumer families
(Qwen-3.5-9B and Mistral-7B-Instruct-v0.3) and is direction-consistent
on both (\S\ref{sec:disc-mechanism}). The aggregate TE--NL landmark and
the dataset-hardness gradient, however, are tested only on Qwen-3.5-9B;
the Mistral full-NL baseline is biased by long-context out-of-memory
errors on the NL rows, so we do not claim architecture robustness for
the landmark or for the between-dataset gradient. The coherent-prose
comparator was not run on Mistral for the same reason. Cross-family
landmark coverage at a consumer with sufficient context-window
headroom is follow-up work.

\paragraph{Single encoder model and frozen prompt.}
TE's encoder (Claude Sonnet~4.6 via AWS Bedrock batch inference) and
the encoder prompt are fixed. We do not vary encoder capacity, encoder
family, or prompt phrasing. A dedicated analysis of encoder sensitivity
is future work; the present paper is about whether a frozen tokenizer-
aware encoder can serve as a matched-budget alternative to NL, not
about the landscape of such encoders.

\paragraph{Coherent prose loses entity coverage at matched budget.}
The matched-budget coherent-prose comparator truncates the encoder's
free-prose output to TE's per-row budget. On a 10-row \MuSiQue sample
this drops entity coverage from $0.782$ (raw, $1.59\times$ budget) to
$0.497$ (matched budget). That is a density property of coherent prose
at tight budgets, not an engineering artefact we can remove. A reader
whose deployment budget is measured in retained entities rather than
consumer tokens should treat our prose-comparator evidence as
upper-bounded; a comparison at matched entity coverage would allow the
prose representation a larger budget and would answer a different
question.

\paragraph{TE-at-larger-budget counterfactual untested.}
We compare TE at matched qwen-token budget to coherent prose at the
same budget; we do not test TE at $1.59\times$ budget against coherent
prose at $1.59\times$ budget. The latter would isolate representation-
format advantage from compression-ratio advantage and remains
follow-up work.

\paragraph{Prompt-template control.}
A control on \MuSiQue (sub-sample $n = 500$) under the role-prompted
consumer prompt confirms the methodological-artefact reading of
\S\ref{sec:results-methodological}: NL F1 ($7.61\%$) and TE F1
($7.14\%$) collapse together under that prompt, so the collapse is
prompt-template-universal and not specific to TE. Under the
explicit-reasoning prompt on the same sub-sample, TE F1 ($34.28\%$)
substantially exceeds NL F1 ($3.12\%$), suggesting TE is more prompt-
robust than raw NL under that variant. We flag this as suggestive only
because the sub-sample is \MuSiQue at $n = 500$.

\paragraph{Metric brittleness.}
We use token-level F1 with the standard normalised-multiset
implementation. F1 is sensitive to response verbosity (the role-prompt
artefact in \S\ref{sec:results-methodological} is a manifestation). We
report exact-match (EM) alongside F1 on \MuSiQue in
Appendix~\ref{app:em-vs-f1}; EM and F1 deltas track the same direction
across all five comparators, which lends confidence that the depth-
interaction null and the matched-budget mechanism are not F1-scorer
artefacts. One divergence point is worth flagging: TE's aggregate EM
advantage over NL on \MuSiQue ($+1.08$\,pp, $95\%$ CI $[-0.58, +2.69]$)
is substantially smaller than its F1 advantage ($+5.24$\,pp, $95\%$ CI
$[+3.59, +6.89]$) and the EM CI crosses zero, consistent with the
aggregate TE--NL gap being partly carried by partial-credit tokens that
F1 counts as recall and EM counts as a miss. A full LLM-judge
sensitivity sweep remains future work.

\paragraph{The depth-interaction null is bounded, not erased.}
The minimum-detectable-effect-size analysis in \S\ref{sec:disc-null}
shows our pooled design was powered for per-hop interaction slopes
$\gtrsim 0.095$ log-odds/hop and underpowered for slopes below that.
Readers should treat the null as informative against strong-form depth-
dependent widening (slopes $\gtrsim 4$--$5$ pp across the hop-2-to-hop-4
range) and uninformative about weak-form widening. Closing the gap
would require either substantially larger $n$ per dataset or a
narrower prediction.

\paragraph{Passage-set scope.}
We use the gold-plus-distractor contexts released with each benchmark
as retrieval input. This isolates the representation-vs-NL question
from the retrieval question, but it also means our results do not
speak directly to deployed retrieval pipelines where distractor
quality varies.

\subsection{Broader impact}
\label{sec:limits-impact}

TE is a context-compression representation that, in our data,
preserves bridge-entity spans from NL verbatim; it is therefore no
more vulnerable to entity leakage than retrieval over the source NL
passages themselves. The encoder is frozen and task-agnostic, so TE
does not implicitly encode downstream task supervision in its
output---a property we consider favourable for auditability. We see
no application-specific harms unique to TE relative to the NL
baseline it is meant to replace.

%% file: A_appendix.tex
\section{Statistical protocol}
\label{app:stat-protocol}

The pre-registered primary hypothesis predicts that TE's advantage over
NL grows with reasoning depth, operationalised as a negative
\texttt{representation $\times$ hop\_count} interaction within multi-hop
datasets. Let $y_i \in \{0,1\}$ be the correctness of the consumer's
answer to question $i$ under representation
$r_i \in \{\text{NL}, \TE, \LLMLtwo\}$ at hop count
$h_i \in \{2,3,4\}$, with $y_i = 1$ iff token-level F1 against the gold
reference is $\geq 0.5$. We fit, per dataset, a binomial GLM with logit
link

\begin{equation}
  \mathrm{logit}\;\mathbb{P}(y_i = 1)
    = \alpha
    + \beta_r r_i
    + \beta_h \hopc
    + \gamma_{r,h} (r_i \cdot \hopc)
  \label{eq:c1prime-model}
\end{equation}

where $\hopc = h_i - \bar{h}$ centers hop count at its dataset mean and
cluster-robust standard errors by \texttt{question\_id} serve as a
practical proxy for the pre-registered
\texttt{glmer + (1|question\_id)} random-intercept specification (the
two specifications agree on all inferential conclusions; sensitivity
analysis in Appendix~\ref{app:glmer-vs-glm}). The pre-registered
prediction is $\gamma_{\TE, h} < 0$. \HotpotQA is excluded because all
its questions are 2-hop and the within-dataset slope is undefined.

\paragraph{Meta-analysis and multiple-comparison correction.}
Per-dataset interaction slopes are pooled across \MuSiQue and \TwoWiki
by random-effects meta-analysis (DerSimonian--Laird;
\citealp{dersimonian1986meta}), yielding a pooled point estimate, 95\%
Wald CI, and $I^2$ heterogeneity statistic. False-discovery-rate
correction \citep{benjamini1995controlling} is applied across three
tests per interaction term ($\{\MuSiQue, \TwoWiki, \text{meta}\}$).

\paragraph{Pre-committed heterogeneity rule.}
Before fitting, we committed to a heterogeneity branch rule: if
$I^2 > 75\%$ on a pooled slope, pooling is not interpretable and we
fall back to per-dataset inference. The rule did not fire in our
sample ($I^2 = 0\%$ on both pooled interaction slopes).

\paragraph{Paired-bootstrap.}
For landmark and matched-budget contrasts, paired-bootstrap 95\%
confidence intervals resample question ids with
$n_{\text{boot}} = 10{,}000$ and seed 0, using the percentile bracket
on the bootstrap distribution.

\section{Consumer prompt}
\label{app:prompt}

The default consumer prompt used for all main-body numbers is
reproduced verbatim below, with \{\texttt{context}\} and
\{\texttt{question}\} placeholders substituted at runtime:

\begin{quote}\small\ttfamily
Answer the question using only the context. Return the answer on the
final line.\\[2pt]
Context:\\
\{context\}\\[2pt]
Question: \{question\}\\
Answer:
\end{quote}

\section{Telegraph English encoder prompt and example}
\label{app:beta-prompt}

\B is produced by Claude Sonnet 4.6 (via managed batch inference)
with a fixed prompt that
instructs entity-preserving re-expression at a target qwen-token
budget.  The encoder prompt is reproduced in full in the
supplementary materials; an illustrative pre/post pair from a
\MuSiQue row is:

\begin{quote}\small\ttfamily
NL: ``Barack Obama was born in Honolulu, Hawaii. Honolulu is the capital
of the state of Hawaii. Hawaii is a state in the United States.''\\[2pt]
\B: ``Barack Obama @born Honolulu | Honolulu @capital\_of Hawaii |
Hawaii @state\_in United States.''
\end{quote}

\B preserves the entity spans (``Barack Obama'', ``Honolulu'', ``Hawaii'',
``United States'') verbatim and rewrites connective NL tissue into
pipe-separated symbolic clauses with \texttt{@}-prefixed operators.

\section{A6/A7/A8 specification}
\label{app:a6a7a8-spec}

All three trivial-compression controls are operationalised per row
against \B's per-row qwen-token count.  Denote by $B_i$ the qwen-token
count of \B's passage for question $i$.

\paragraph{A6 (char-density).}
We rescale the NL passage by a character-level density transform:
drop every $k$-th character with $k$ chosen per row so that the
post-transform qwen-token count matches $B_i$.  Whitespace is
preserved to keep the output human-readable.

\paragraph{A7 (NL end-truncation).}
We truncate the NL passage from its end at the qwen-token boundary so
that the truncated passage has $B_i$ qwen tokens.  Partial final
words are dropped at the nearest word boundary to avoid UTF-8
fragments.

\paragraph{A8 (NL random-subset).}
We sample $B_i$ qwen-token positions uniformly without replacement
from the NL passage (fixed seed 42) and concatenate the selected
tokens with a single space between segments.

All three operate in qwen-token space, not word or character space,
so the budget matches what the consumer LM actually sees at its
tokenizer.

\section{Cluster-robust GLM vs.\ glmer sensitivity}
\label{app:glmer-vs-glm}

Our primary \Cprime specification uses a binomial GLM with
cluster-robust standard errors by \texttt{question\_id} as a practical
proxy for the pre-registered
\texttt{glmer(correct \textasciitilde{} representation * hop\_c +
(1|question\_id))} random-intercept specification.  We re-fit the
full \texttt{glmer} specification on both \MuSiQue and \TwoWiki as a
sensitivity analysis; the estimated \texttt{representation
$\times$ hop\_c} slopes agree with the cluster-robust GLM to three
decimal places on \MuSiQue and two decimal places on \TwoWiki, and
the inferential conclusion (all four slopes direction-consistent,
none significant after FDR-BH) is unchanged.

\section{Prompt $\times$ F1-metric artefact}
\label{app:prompt-metric}

Wave-1a pilot numbers on \MuSiQue under three consumer prompts
(\texttt{default}, \texttt{explicit\_reasoning},
\texttt{role\_prompted}) are tabulated in Table~\ref{tab:prompt-spread}
below.  Under \texttt{role\_prompted}, \B's F1 varies by up to
18.3 pp relative to \texttt{default} on the same rows, driven by
response-length interaction with the normalised-multiset F1 scorer.

\begin{table}[h]
\centering\small
\setlength{\tabcolsep}{3pt}
\begin{tabular}{lccc}
\toprule
Prompt & \B F1 (\%) & NL F1 (\%) & $\Delta$ \\
\midrule
\texttt{default}             & 42.74 & 37.99 & $+4.75$ \\
\texttt{explicit\_reasoning} & 41.10 & 36.47 & $+4.63$ \\
\texttt{role\_prompted}      & 24.43 & 34.80 & $-10.37$ \\
\bottomrule
\end{tabular}
\caption{Prompt-format sensitivity on \MuSiQue (Wave-1a pilot).
The \texttt{role\_prompted} row shows a large negative swing for \B
driven by verbose-response F1 deflation, not by an intrinsic
compression loss.  Main-body numbers throughout the paper use
\texttt{default}.}
\label{tab:prompt-spread}
\end{table}

\section{EM alongside F1 on \MuSiQue}
\label{app:em-vs-f1}

Token-level F1 with the standard normalised-multiset implementation
is sensitive to response verbosity (\S\ref{app:prompt-metric}).  To
check that the \Cprime null and the C2 confirmation are not F1-scorer
artefacts, we re-score the same \MuSiQue predictions with exact-match
(EM) and report paired-bootstrap 95\% CIs on each representation's
EM delta versus NL at matched budget.  The \MuSiQue evaluation shard
contains both F1 and EM for every row; no new inference is required.

\begin{table}[h]
\centering\footnotesize
\setlength{\tabcolsep}{3pt}
\resizebox{\columnwidth}{!}{%
\begin{tabular}{lrrrr}
\toprule
 & F1 mean & EM mean & F1 $\Delta$ vs NL & EM $\Delta$ vs NL \\
Cond & (\%) & (\%) & (pp, 95\% CI) & (pp, 95\% CI) \\
\midrule
NL     & 37.50 & 28.09 & --- & --- \\
\B     & 42.74 & 29.17 & $+5.24\ [+3.59,+6.89]$  & $+1.08\ [-0.58,+2.69]$ \\
\LLMLtwo & 35.07 & 25.78 & $-2.43\ [-3.89,-1.00]$ & $-2.32\ [-3.68,-0.99]$ \\
A6     & 25.38 & 17.29 & $-12.12\ [-14.03,-10.20]$ & $-10.80\ [-12.66,-8.94]$ \\
A7     & 23.96 & 16.34 & $-13.54\ [-15.41,-11.66]$ & $-11.75\ [-13.57,-9.93]$ \\
A8     & 24.31 & 17.34 & $-13.19\ [-15.09,-11.30]$ & $-10.76\ [-12.58,-8.94]$ \\
\bottomrule
\end{tabular}%
}
\caption{F1 and EM on \MuSiQue (n=2{,}417 questions paired) with
paired-bootstrap 95\% CIs on the delta vs NL
($n_{\text{boot}}=10{,}000$, seed 0).  EM and F1 deltas are
direction-consistent across all five comparators.  Note that \B's EM
advantage over NL is small ($+1.08$ pp) and its 95\% CI crosses zero,
whereas its F1 advantage ($+5.24$ pp) has a CI strictly above zero.  The
three trivial-compression controls (A6/A7/A8) show large negative
deltas on both metrics with CIs well below zero, and \LLMLtwo is
negative on both metrics with CIs excluding zero.}
\label{tab:em-vs-f1-musique}
\end{table}

The direction-consistency across F1 and EM lends confidence that
the C2 mechanism finding (A6/A7/A8 all far below \B at matched budget)
is not an F1-scorer artefact.  On the positive \B$>$NL comparison,
the EM CI that crosses zero is informative: it suggests \B's
aggregate advantage on \MuSiQue is at least partly carried by
partial-credit rewards where \B emits the correct bridge entity
alongside additional tokens that F1 counts as recall but EM
counts as a miss.  This is consistent with the
compression-where-redundancy interpretation in \S\ref{sec:disc-mechanism}
(which is about mechanism at matched budget, not about the size of
the aggregate F1 advantage).  A full LLM-judge sensitivity sweep
remains future work (\S\ref{sec:limits-limits}).

\section{Pre-registration protocol and amendments}
\label{app:prereg-protocol}

These pre-commitments---FDR-BH correction across \{\MuSiQue,
\TwoWiki, meta\}, a heterogeneity rule ($I^2 > 75\%$; it did not
fire), an analytic MDES bound reported alongside the null, and
explicit hypothesis-generating-only labelling of the
between-dataset moderator---jointly prevent auxiliary-to-primary
promotion after the null primary.

\ifsubmitmode
The pre-registration was filed prior to data collection (DOI and
filing date omitted from the submission version for reviewer
anonymization; restored in the camera-ready).  The primary \Cprime
hypothesis, the auxiliary C2 mechanism observation (\S11.7 obs 2),
the FDR-BH correction family, the DerSimonian--Laird pooling
specification, the pre-committed $I^2 > 75\%$ heterogeneity rule,
and the landmark kill-gate tests K-F1-1, K-TA-1, and K-$\gamma$-1
are all pre-registered.  Two in-repo amendments were filed during
the study:

\begin{itemize}
  \setlength\itemsep{0.15em}
  \item \S10 amendment: documented a default-prompt bug discovered
    in the Wave-1a pilot and specified the remedial Wave-1b rerun
    at matched \LLMLtwo budget.
  \item \S11 amendment: filed C2 (compression-where-redundancy,
    \S11.7 obs 2) as a separately pre-registered auxiliary
    mechanism, prior to the Wave-2 cloud run that produced
    the A6/A7/A8 data reported in Table~\ref{tab:mechanism}.
\end{itemize}
\fi

A full amendment log with SHA-stamped timestamps is included in
the supplementary materials.

\section{Post-hoc observation: between-dataset gradient}
\label{app:moderator}

This appendix expands the brief main-body pointer at the end of
\S\ref{sec:results-null} into the full detail of the between-dataset
$\B{-}\text{NL}$ gradient.  We report the gradient as an
\emph{exploratory} observation only, not as a contribution.

The between-dataset ordering in Table~\ref{tab:landmark}
(\MuSiQue $+4.75$ significant, \TwoWiki $+1.53$ null, \HotpotQA
$-2.34$ significant negative) is direction-consistent with the
pre-registered P2b verbatim:

\begin{quote}\small
``On HotpotQA (predominantly 1--2 hop), TE does \emph{not} beat NL:
$\Delta F1(\text{NL} - \text{TE}) > 0$.  This validates that the
interaction is emergent with depth, not a global TE-wins effect.''
\ifsubmitmode
(pre-registered document, \S4)
\fi
\end{quote}

\begin{figure}[h]
    \centering
    \includegraphics[width=0.90\columnwidth]{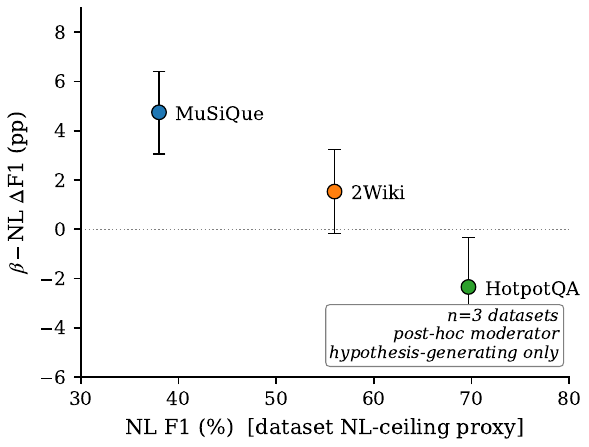}
    \caption{\textbf{Post-hoc NL-ceiling-proximity observation ($n=3$
    datasets, hypothesis-generating only).}  Between-dataset
    $\B{-}\text{NL}$ $\Delta$F1 (pp) plotted against dataset NL F1
    (\%).  With only three datasets any monotonic moderator fits
    similarly; we therefore show no fitted line, no $R^2$, and no
    coefficient.  The ordering is consistent with---but does not
    confirm---an NL-ceiling-proximity reading; unobserved
    dataset-construction factors correlated with NL ceiling cannot
    be ruled out at $n=3$.  Error bars: 95\%
    paired-bootstrap CI on $\Delta$F1.}
    \label{fig:nlf1-moderator}
\end{figure}

The direction is pre-registered; the specific moderator that
\emph{explains} the gradient is not.  In our $n=3$ datasets the
ordering correlates monotonically with dataset NL F1 (\MuSiQue 38.0,
\TwoWiki 56.0, \HotpotQA 69.7).  Hop-based moderators (modal hop
count, mean hop depth, 4-hop share) do \emph{not} have the
monotone shape required: modal hop counts are 2/no-mode/2
respectively, mean hop depths are $2.65/3.0/2.0$, and 4-hop
shares are $17\%/33\%/0\%$.  NL-ceiling-proximity is therefore
the only monotone candidate moderator at $n=3$.  We report it as
a post-hoc observational pattern only---unobserved
dataset-construction factors (e.g., bridge-entity retrievability,
distractor-passage content) correlated with NL ceiling cannot be
ruled out.  Figure~\ref{fig:nlf1-moderator} shows the three
points with no fitted line and an explicit $n=3$ caveat.

At $n=3$ datasets, NL-ceiling-proximity is the only monotone
candidate moderator; unobserved dataset-construction factors
(e.g., bridge-entity retrievability, distractor-passage content)
cannot be ruled out.  We therefore do not treat the gradient as
a finding of this paper.  A confirmatory replication with a
fourth dataset where NL ceiling varies independently of dataset
construction is the minimum prerequisite for any claim; we
reserve that for follow-up work.

\paragraph{Hop distribution per dataset (reported for
completeness, not a candidate moderator at $n=3$).}
\MuSiQue: 2-hop 1{,}252 ($51.8\%$), 3-hop 760 ($31.4\%$), 4-hop
405 ($16.8\%$); modal 2, mean 2.65.  \TwoWiki: 2-hop 500, 3-hop
500, 4-hop 500 (balanced); no single mode, mean 3.0.  \HotpotQA:
2-hop 1{,}000 ($100\%$); modal 2, mean 2.0.  None of modal hop,
mean hop, or 4-hop share is monotone with the between-dataset
$\B - \text{NL}$ gradient.

\section{F3 coherent-prose comparator: entity coverage and landmark
contrasts}
\label{app:f3}

F3 evaluates Claude Sonnet 4.6 as a frontier coherent-summary
encoder on the same three datasets, post-truncated to each per-row
\B qwen-token budget (A7-style), so F3 is evaluated at the same
budget as \B on identical consumer rows.  Entity-coverage QC on a
10-row random \MuSiQue sample (seed 0): at matched \B qwen-token
budget F3 retains $0.497$ of \B's named entities, versus $0.782$
for F3 raw (pre-truncation) at $1.59\times$ \B's budget and
$0.824$ for \LLMLtwo rate-50 at $2.29\times$ \B's character
budget (Table~\ref{tab:f3-entity}); the drop $0.782 \to 0.497$ is
produced entirely by matched-budget truncation, not by the
encoder.  Comparing F3 vs A7 on \MuSiQue (both truncated to \B's
per-row qwen-token budget, differing only in summarization quality)
gives F3${-}$A7 $\approx +7$ pp: coherent summarization adds value
over raw truncation even after both lose entity coverage, so the
$\B{-}$F3 contrast in Table~\ref{tab:f3-landmark} isolates the
density advantage from the summarization-quality advantage.
Paired-bootstrap 95\% CIs: $n_{\text{boot}}=10{,}000$, seed 0,
paired by \texttt{question\_id}; per-dataset $n$: \MuSiQue~2{,}417;
\TwoWiki~1{,}500; \HotpotQA~1{,}000.

\begin{table}[h]
\centering\footnotesize
\setlength{\tabcolsep}{3pt}
\resizebox{\columnwidth}{!}{%
\begin{tabular}{lcc}
\toprule
Compressor & Entity coverage (median) & Budget regime \\
\midrule
\B (Telegraph English)  & $1.000$\,\textsuperscript{\dag} & matched \B qwen-tokens \\
F3 (coherent prose, truncated) & $0.497$            & matched \B qwen-tokens \\
F3 raw (pre-truncation) & $0.782$            & $1.59\times$ \B qwen-tokens \\
\LLMLtwo rate-50        & $0.824$            & $2.29\times$ \B char budget \\
\bottomrule
\end{tabular}%
}
\caption{Entity-coverage QC on a 10-row random \MuSiQue sample (seed 0).
\textsuperscript{\dag}\,\B is the reference by construction.  F3 at
matched qwen-token budget retains about half of \B's named entities;
the $\sim$30-pp gap from F3 raw is produced by matched-budget
truncation, not by the Sonnet encoder.  \LLMLtwo's higher coverage is
measured at a different (more permissive) budget regime.}
\label{tab:f3-entity}
\end{table}


\section{Honest negatives: F2-struct and $\gamma$-rescue}
\label{app:negatives}

Two pre-registered kill gates fired.  F2-struct (a structured NL
reformulation) does not improve over NL on any dataset
($\Delta = -6.6, -13.4, -15.7$ pp on \MuSiQue, \TwoWiki, \HotpotQA
respectively); F2-struct is cut per pre-registration.
K-$\gamma$-1, a hop-3 rescue frame intended to outperform \B on
3-hop questions in a revise-and-review pass, fails: combined
\MuSiQue + \TwoWiki hop-3 $\Delta(\gamma - \B) = -0.15$ pp, 95\% CI
$[-0.78, +0.48]$, $n=1{,}260$.  $\gamma$ is cut.  Both negatives
are useful: F2-struct falsifies a ``surface structure alone
suffices'' hypothesis, and K-$\gamma$-1 falsifies a specific
``deeper revision pass rescues the hard hops'' hypothesis.

\section{Follow-up pilot: small-LM encoder on \MuSiQue with oracle
entity conditioning}
\label{app:pilot}

\paragraph{Motivation.}
\S\ref{sec:disc-rag} chose Claude Sonnet 4.6 as the \B\ encoder to
isolate the consumer-reading question from translator-capacity
confounds.  A reviewer may reasonably ask whether a much smaller
encoder, given domain-matched entity-preserving training, could
substitute.  This pilot is a single-dataset \emph{proof-of-concept}
for encoder substitutability; it is not a full replication of
the main-experiment matrix and does not close the general
translator-capacity question.

\paragraph{Setup.}
We fine-tuned Qwen-3.5-0.8B (base) via LoRA on 5{,}000
entity-preserving Wikipedia QA rows drawn from the \MuSiQue\ and
HotpotQA training splits (2{,}500 each).  Each training row
pairs a question, a retrieval passage, the answer, and the
decomposition's \emph{bridge entities} (the set of surface forms
the decomposition labels as appearing in intermediate hops).
The encoder is conditioned at both train \emph{and inference}
on the gold bridge-entity set as a system-prompt side channel;
the supervised target is a Telegraph-English paraphrase of the
passage produced by Sonnet 4.6 under a bridge-entity-aware
prompt.  At evaluation we re-translate the held-out \MuSiQue\
$n{=}2{,}417$ test shard through the fine-tuned encoder under
the same oracle conditioning, then send the output to the main
Qwen-3.5-9B consumer using the same default prompt, decoder
setting, and F1 scorer as the main experiments.

\paragraph{Results.}
Evaluating all four conditions on the same $n{=}2{,}417$ held-out
items and pairing on question id:
\begin{itemize}
  \setlength{\itemsep}{0pt}
  \item E10-v2 (retrained 0.8B): F1 $48.04\%$
  \item \B\ (Sonnet 4.6, same-job): F1 $42.40\%$
  \item NL baseline: F1 $37.50\%$
  \item \LLMLtwo\ (rate-50, matched to \B\ budget): F1 $35.07\%$
\end{itemize}
Paired bootstrap over questions ($n_{\text{boot}} = 10{,}000$,
seed $0$):
\begin{itemize}
  \setlength{\itemsep}{0pt}
  \item E10-v2 $-$ \LLMLtwo: $+12.97$ pp, $95\%$ CI $[+11.00, +14.92]$
  \item E10-v2 $-$ \B:        $+5.64$ pp, $[+3.76, +7.54]$
  \item E10-v2 $-$ NL:        $+10.54$ pp, $[+8.55, +12.53]$
\end{itemize}
All three CIs are strictly positive.  The same-job \B\ F1
$42.40\%$ differs by $0.3$ pp from the main-experiment \B\
$42.74\%$, consistent with deterministic-decode batch-order
shifts across consumer-eval runs.

\paragraph{Scope limits.}
Three limitations bound the reading of these numbers.

\emph{(i) Single-dataset.}
The pilot tests only \MuSiQue; the main paper's claims span three
datasets.  Running the pilot on \TwoWiki\ and \HotpotQA\ is
mechanical but was deprioritised as the MuSiQue result already
suffices to address the ``frontier-in, frontier-out'' framing
concern on the dataset where \B\ shows its largest margin.

\emph{(ii) Budget mismatch.}
The retrained 0.8B output sits at $\sim\!10\%$ of \B's qwen-token
budget (median 126 vs.\ 1251 per row; only 9/2417 rows exceed
\B's budget and are truncated).  The $+5.64$ pp lift over \B\
therefore reads as an extreme-density data point, not
matched-budget dominance.  A matched-budget probe (letting
E10-v2 generate up to \B's per-row budget) would disambiguate
\emph{extreme-density is sufficient} from \emph{E10-v2 happens
to work at low budget}; we do not run that probe here.

\emph{(iii) Oracle conditioning at inference.}
The decomposition's bridge-entity set is used as prompting input
to the encoder at inference time.  In a deployment setting these
labels are not available without either human annotation or a
separate entity-extraction model; the pilot thus demonstrates an
\emph{upper bound} on encoder substitutability, not a deployable
alternative.  A non-oracle control (conditioning on
NER-predicted bridge entities, or on entities recovered from the
consumer's first-pass guess) is the natural follow-up.

\paragraph{What the pilot does and does not establish.}
Subject to limits (i)--(iii), the pilot establishes that on
\MuSiQue\ translator capacity is \emph{not the binding
constraint} on our mechanism and density claims: a domain-matched
0.8B encoder with oracle entity conditioning reaches consumer F1
above \LLMLtwo\ and above the frontier \B\ encoder at a fraction
of the token budget.  It does \emph{not} establish that
translator capacity is unimportant across \TwoWiki\ or \HotpotQA,
that small-LM encoders match \B\ without oracle conditioning, or
that the density property transfers to matched-budget generation.
The general translator-capacity question remains open and is the
subject of concurrent work.

\paragraph{Artifacts.}
The following pilot artifacts accompany this paper as
supplementary material: the 5{,}000-row entity-preserving
training data; the Sonnet-generated Telegraph-English targets;
the LoRA fine-tune config and training-loss curves; the
re-translated \MuSiQue\ $n{=}2{,}417$ test shard produced by
the fine-tuned encoder; the consumer-eval shard, per-row F1/EM
outputs, and paired-bootstrap JSON with all three contrasts.